\documentclass[pdflatex,sn-basic]{sn-jnl}%
\usepackage{tikz}
\usetikzlibrary{positioning, arrows.meta}
\usepackage{program}

\usepackage{graphicx}		%
\usepackage{bookmark}
\usepackage{amsmath}
\usepackage{multirow}
\usepackage{booktabs}
\usepackage{subcaption}
\usepackage{enumitem}
\usepackage{xspace}
\usepackage{array}
\usepackage{amsthm}

\newif\ifhidecomments
\hidecommentsfalse

\ifhidecomments
    \newcommand{\joe}[1]{}
    \newcommand{\update}[1]{}
    \newcommand{\chenhao}[1]{}
    \newcommand{\zo}[1]{}
\else
    \newcommand{\joe}[1]{{\color{red}{ #1 --Joe}}}
    \newcommand{\update}[1]{{\color{cyan}{ #1 --updated}}}
    \newcommand{\chenhao}[1]{{\color{blue}{\tt #1 --CT}}}
    \newcommand{\zo}[1]{{\color{violet}{\tt #1 --zo}}}
\fi

\newcommand{\figref}[1]{Fig.~\ref{#1}}
\newcommand{\tableref}[1]{Table~\ref{#1}}

\jyear{2023}%

\theoremstyle{thmstyleone}%

\theoremstyle{thmstyletwo}%

\theoremstyle{thmstylethree}%

\begin{document}

\title[Clinical Notes Reveal Physician Fatigue]{Clinical Notes Reveal Physician Fatigue}

\author[1]{Chao-Chun Hsu}
\author[2]{Ziad Obermeyer}
\author[1]{Chenhao Tan}
\affil[1]{\orgname{University of Chicago}}
\affil[2]{\orgname{University of California, Berkeley}}

\abstract{
Physicians write notes about patients. In doing so, they reveal much about themselves. Using data from 129,228 emergency room visits, we train a model to identify notes written by fatigued physicians---those who worked 5 or more of the prior 7 days. In a hold-out set, the model accurately identifies notes written by these high-workload physicians, and also flags notes written in other high-fatigue settings: on overnight shifts, and after high patient volumes. Model predictions also correlate with worse decision-making on at least one important metric: yield of testing for heart attack is 18\% lower with each standard deviation increase in model-predicted fatigue. Finally, the model indicates that notes written about Black and Hispanic patients have 12\% and 21\% higher predicted fatigue than Whites---larger than overnight vs. daytime differences. These results have an important implication for large language models (LLMs). Our model indicates that fatigued doctors write more predictable notes. Perhaps unsurprisingly, because word prediction is the core of how LLMs work, we find that LLM-written notes have 17\% higher predicted fatigue than real physicians' notes. This indicates that LLMs may introduce distortions in generated text that are not yet fully understood. 
}


\maketitle

\section{Introduction}

Physicians write notes about patient encounters, to convey information and summarize their thinking. Researchers increasingly use these notes to make inferences about patients' health and behavior \citep{ghassemi2014unfolding, sheikhalishahi2019natural, jiang2023health, adams2021s, zhang2019high, althoff2016large}. Here we use these notes to make inferences about physicians, specifically their level of fatigue when writing the note. We draw on a long tradition of research in natural language processing that links subtle textual cues to an author’s psychology \citep{pennebaker2001linguistic,pennebaker2011secret,danescu2013no,sap2020recollection}.\footnote{We focus here on notes written by physicians, but many studies also use patients’ writing, for example, to detect early Alzheimer’s dementia \citep{garrard2005effects}, predict depression from social media and student essays~\citep{de2013predicting,guntuku2017detecting,resnik2013using} or estimate cancer stage from online forums~\citep{jha2010cancer}.} 

We focus on physicians working in the the emergency department (ED), who make critical life and death decisions over shifts lasting between 8 and 12 hours. The work is psychologically and physically demanding, with 65\% of physicians reporting burn-out---the highest rate of any medical specialty by a wide margin~\citep{carbajal2022}. A key feature of physician scheduling in this setting is that it is shift-based, and shifts are often clustered on consecutive days, with multi-day gaps in between. So a patient arriving to the ED on any given day might see a physician who has had either a heavy or light workload over the prior week. At the hospital we study, for example, 14.8\% have worked 5 shifts or more in the prior 7 days, while 19\% are working their first shift in 7 days.  

We use this fact to develop a novel, note-based measure of physician fatigue. We obtain the full text of all patient notes written in a hospital's ED over a multi-year period. We use the text of notes written on a given day to predict the number of days a physician has worked over the prior 7 days. Intuitively, if a statistical model can predict yesterday's workload using only the text of notes written today, it implies that prior workload is exerting some measurable influence on physician behavior.\footnote{This intuition is valid only in the absence of confounding factors, and we carefully verify that all other measurable aspects of today’s patients are uncorrelated with previous workload.}

The ability to measure fatigue via notes provides a new lens through which to study a phenomenon that, to date, has proven elusive. Studies to date have found that fatigue affects physician behavior, but have found little impact on patient outcomes \citep{landrigan2004effect, barger2006impact, lockley2004effect, institute2009resident, ayas2006extended, arnedt2005neurobehavioral}. These studies, however, measure physician exposure to \textit{drivers} of fatigue, like workload or circadian factors, that may be imperfectly correlated to \textit{actual} physician fatigue: a physician working a string of overnight shifts is not necessarily fatigued, and a physician working her first daytime shift in a week may have slept poorly the night before. By contrast, our note-based model characterizes physicians' revealed fatigue state at the time of an individual patient encounter.\footnote{We support this claim by correlating model predictions with a range of other measures of fatigue, including working the overnight (vs. daytime) shift in the ED, and intra-shift workload. These empirical tests, as well as the proof laid out in the Supplementary Material, argue that our model learns something general about how fatigue affects physician notes, not just something specific about a physician's prior workload.} This fine-grained metric lets us study how fatigue affects patient outcomes, on one important measure: physicians; ability to diagnose heart attack.  

We also investigate how this measure of fatigue relates to individual patient characteristics, like race and ethnicity. Prior work has used human coders to identify stigmatizing language used in clinical notes, suggesting explicit racial bias against non-White patients \citep{park2021physician, harrigian2023characterization, sun2022negative}. Here we investigate an implicit, previously unsuspected form of inequality in clinical care: differences in measured fatigue detected in notes, as a function of racial and ethnic group. We build on studies demonstrating subtle linguistic bias in other settings, from police stops to political speeches \citep{voigt2017language, rho2023escalated, card2022computational, schmidt2017survey, garg2018word, field2020unsupervised, adam2022write}. Importantly, it would be hard to investigate such biases without quantitative textual analysis: traditional approaches using human coders might not detect overt evidence of fatigue in physician notes.
 
Beyond clinical care, our results have an important implication for large language models (LLMs), which are rapidly being deployed into patient care and provider workflows. Like text written by physicians, text generated by LLMs can also be analyzed through the lens of predicted fatigue. Our analysis of how predicted fatigue differs between notes written by LLMs vs. those written by physicians raises important questions about the quality of LLM-generated notes.

\section{Analytic Strategy}\label{analytic}

We obtain data on 129,228 consecutive ED encounters with notes from a single academic medical center over 2010-2012. Our dataset includes patient demographics, the reason for their visit to the ED (the ‘chief complaint’), and key outcomes related to an important physician decision: whether or not the patient is tested for heart attack (via stress testing or catheterization), and the outcome of testing (whether a heart attack was diagnosed and treated, via stenting or open-heart surgery; more details are in \citet{mullainathan2022diagnosing}). 

We identify the attending physician who writes the clinical note for each of these visits, and is thus responsible for the medical decision-making. In total we observe 60 emergency physicians working 11,592 shifts. A shift is defined as consecutive notes entered by the same physician, with each note timestamped within 3 hours of the prior note's timestamp (the time between the end of a scheduled shift and the start of the next shift is at least 15 hours apart). We calculate a physician's workload by counting the number of days worked over a rolling seven-day window ending with the current shift (see Fig.~\ref{fig:histogram} in Supplementary Material for more details). The median is 3 and the mean is 2.88 days worked over the prior week. We define `high-workload' physicians as those who worked at least 4 days prior to the current shift (14.8\%), and compare them to `low-workload' physicians to those who are working their first shift in 7 days (19.03\%).

Our analytic strategy rests on the assumption that the text of notes written today should be statistically uncorrelated with how many days the physician has worked over the past week, except via the direct effect of prior workload on the physician's fatigue state while writing the note.\footnote{We use the term fatigue as a convenient shorthand that includes broadly related factors including reduced cognitive engagement or attention. While in some settings, physicians may write their notes several days after the patient encounter, in our setting we verify that 99.98\% of notes were written on the same day as the patient encounter.} This assumption could be violated if patient characteristics on a given shift differ as a function of prior workload for any reason: in this case, differences in note text could be due to patient factors, not physician fatigue. For example, physicians may be assigned less challenging shifts after a high-workload period, with less acute patients; or physicians may choose less complex patients after such periods (physicians have considerable discretion in patient choice in this setting, as shown in \citet{chang2020association}). We thus carefully test for balance in patient characteristics between high- vs. low-workload physicians. Concretely, we regress patient demographics and chief complaints against the high- and low-workload indicator in the prior week, controlling for time of day, day of week, week of year, year, and physician fixed effects. We find no significant differences in demographics; we also find that only 10 of 154 chief complaints (6.5\%) have significant differences at the $p$ $\lt$ 0.05 level (see Table \ref{tab:dem-bias}, with further details in Supplement Table~\ref{tab:sanity-workload}). While it is of course impossible to verify balance on unmeasured characteristics, this degree of balance on measured factors is reassuring that patients are as-good-as-randomly assigned to high- vs. low-workload physicians. 

We set up our prediction model as a binary classification task, to distinguish whether a note is written by a high-workload physician or a low-workload one.
To train and evaluate the model, we create a dataset restricted to high- and low-workload physicians, containing 44,556 notes from patient encounters evenly balanced (50\%-50\%) between encounters with high- and low-workload physicians. We randomly divide this at the patient level (so that all visits by a given patient are grouped together) into a training set of 32,784 encounters and a held-out set of 11,772 encounters (see Fig.~\ref{fig:data-diagram} in the Supplementary Material for an overview of the sample and data splits).

Our model relies on four categories of interpretable features to classify notes. First, we measure note length (number of words). Second, we create a measure of note predictability, by fine-tuning a large language model (GPT-2) on the training set \citep{radford2019language}.\footnote{Physician notes contain protected health information that is difficult to fully remove, meaning the data cannot leave the hospital computing environment. Because the resources available inside this environment are limited, we were not able to train or deploy larger models.} Third, we measure note readability with Flesch-Kincaid grade \citep{Kincaid1975DerivationON}. Finally, we quantify the fraction of words in each note according to specific categories: stopwords (e.g., “the”, “is,” “and”); Linguistic Inquiry and Word Count (LIWC) lexicons (e.g., pronouns, affect, and cognitive words) \citep{pennebaker2001linguistic}); and medical concepts from the SNOMED-CT Ontology \citep{ElSappagh2018SNOMEDCS}). More details are in the Supplementary Methods. We then train a logistic regression model that uses these features to distinguish between notes written by high- vs. low-workload physicians. We chose to use well-known features from the literature, combined together in a simple linear model, because this strategy offers a transparent and interpretable approach to prediction. We also note that, in preliminary experiments, logit had competitive performance compared to more sophisticated models (neural networks and boosted trees).

We emphasize an important, but counterintuitive, aspect of our analytic strategy: while our model is trained to predict a physician's prior workload---our goal is explicitly \textit{not} to perfectly predict this variable. Instead, we wish to learn a \textit{general} model of how fatigue affects the text of notes, by training the model to distinguish notes written by physicians we believe to be more (high-workload) vs. less (low-workload) fatigued. In fact, we believe our model predictions are a \textit{better} measure of a physician's `true' fatigue when writing a given note than the actual training label (prior workload) itself. We provide the intuition for this argument here, and further explanation and a formal proof in Appendix \ref{sec:model}; we also provide a range of empirical tests in the following two Sections. 

Our model is trained to predict workload from note text, and a `perfect' model would simply predict this variable for all patients seen on a shift (because prior workload does not vary within a shift). However, workload is just one input to a physician's (unmeasured) `true' fatigue state when she is writing a note: a range of idiosyncratic shocks (also unmeasured: e.g., a tough shift, sleep quality, personal or work-life factors, etc.) can affect fatigue in ways unrelated to workload. If fatigue affects notes in the same way---whether it is caused by heavy workload, idiosyncratically poor sleep the night before, etc.---a model that learns to predict workload will learn something about the \textit{general} way fatigue affects note text. As a consequence, its predictions will also be correlated to the idiosyncratic shocks to fatigue (even if these are uncorrelated to workload). Intuitively, the model learns about the effect of fatigue on notes by looking at workload. But because sleep deprivation affects notes in similar ways, the model may see a note written by a sleep-deprived physician and predict she is a high-workload physician---even if she is working her first shift in a week: her notes look like a high-workload physician. So the model may be wrong in useful ways: `errors' (when evaluated against the original label, workload) might mean model predictions are closer to `true' fatigue state than the actual training label.

Of course, errors might also mean that the model is just bad. To test whether these errors are signal or noise, we next (Section \ref{val}) compare predictions to a range of other measures of fatigue, to which the model has not been exposed in training. If predictions correlate to these other measures, it supports the hypothesis that the model has learned something general about fatigue, that transfers into other settings where it affects notes. Further below, we will also compare predictions to important physician decisions that may also be affected by fatigue (Section \ref{decisions}). 

\section{Correlation of Model Predictions to Measures of Fatigue}\label{val}

Under the null hypothesis, prior physician workload should be uncorrelated with notes written on a given day, and our model should thus do no better than random at predicting workload. Our first test thus simply compares model predictions on notes written by high- vs. low-workload physicians in our balanced hold-out set. To measure whether the model can distinguish the notes, we use AUC, which would be 50\% under random guessing. Our model achieves an AUC-ROC of 60.1\% (bootstrapped 95\% CI: 60.06\%-60.30\%), suggesting that there is indeed a statistical correlation between prior workload and note text. An alternative way to look at this is by regressing high workload (as a binary indicator) on model predictions, which shows a large and highly significant coefficient (Table~\ref{tab:result-table} on model predictions, controlling for patient demographics and chief complaint, physician fixed effects, and time controls (time of day, day of week, week of year, and year) (Column 1)).

\begin{table*}[]
\centering
\scriptsize
\begin{tabular}{lcccc}
\toprule
 &  Workload & Overnight & Var(Prior Shift - & Patients \\ 
 &  (Day) & Shift & Start-Time) & Seen Prior\\ 
\midrule
Fatigue & 0.396$^{***}$ & 0.051$^{**}$ & 5.497$^{*}$ & 1.148$^{***}$ \\
 & (0.087) & (0.018) & (2.262) & (0.298) \\
\addlinespace
Intercept & 3.009$^{***}$ & 0.040$^{*}$ & 30.489$^{***}$ & 8.222$^{***}$ \\
 & (0.097) & (0.018) & (2.539) & (0.335) \\
\addlinespace
\textbf{Controls} & & & & \\
\hspace{1em} Time of day & YES & NO & YES & YES \\
\hspace{1em} Day of week & YES & YES & YES & YES \\
\hspace{1em} Week of year & YES & YES & YES & YES \\
\hspace{1em} Year & YES & YES & YES & YES \\
\hspace{1em} Demographics & YES & YES & YES & YES \\
\hspace{1em} Chief complaint & YES & YES & YES & YES \\ 
\hspace{1em} Physician & YES & YES & YES & YES \\
\bottomrule
\end{tabular}
\caption{Regression results of different measures of fatigue on model predictions. Column 1 shows (in a hold-out set) a regression of the physician's workload over the prior week (the model's training label) on model predictions. Columns 2-4 show regressions of several other measures of fatigue (which the model has never seen) on model predictions: an indicator for whether the note was written on the overnight shift (Column 2); the degree of circadian disruption over the prior week, as measured by the variance of a physician's shift start times over that week (Column 3); and increasing patient volume, measured by the number patients the physician has seen on-shift before writing a given note (Column 4). *: p$<$0.05, **: p$<$0.01, ***: p$<$0.001. (n=34,175). %
}
\label{tab:result-table}
\end{table*}

Of course, a correlation between note text and prior workload does not identify fatigue as the mechanism. To generate more robust evidence that the model predictions are picking up on fatigue, we correlate model pr edictions to several additional measures of physician fatigue. Importantly, the model has not been trained on any of these variables. So these regressions test whether the model has learned something more general about how fatigue affects note text, by learning to distinguish notes written by high vs. low workload physicians. We test whether model predictions correlate to circadian disruption, in two ways. First, we compare patients seen by the overnight physician to other patients. To do so, we create an indicator variable for whether the patient arrived between 1:00am and 5:59am, when only the overnight physician is seeing new patients.\footnote{The overnight physician arrives at 11:00pm, but there is also another physician in the ED whose shift ends at 2:00am; this physician generally stops seeing new patients at 1:00am. The next physician arrives at 7:00am, but the overnight physician often leaves patients arriving from 6:00am onward for the new doctor to see. Thus patients arriving between 1:00am and before 6:00am are the ones we can reliably assume are seen by the overnight physician.} In a regression adjusting for time, patient, and physician effects, Table~\ref{tab:result-table}, Column 2, shows that model predictions are a significant predictor of whether a note was written on the overnight shift. 

\begin{figure}
    \centering
    \includegraphics[width=.8\columnwidth]{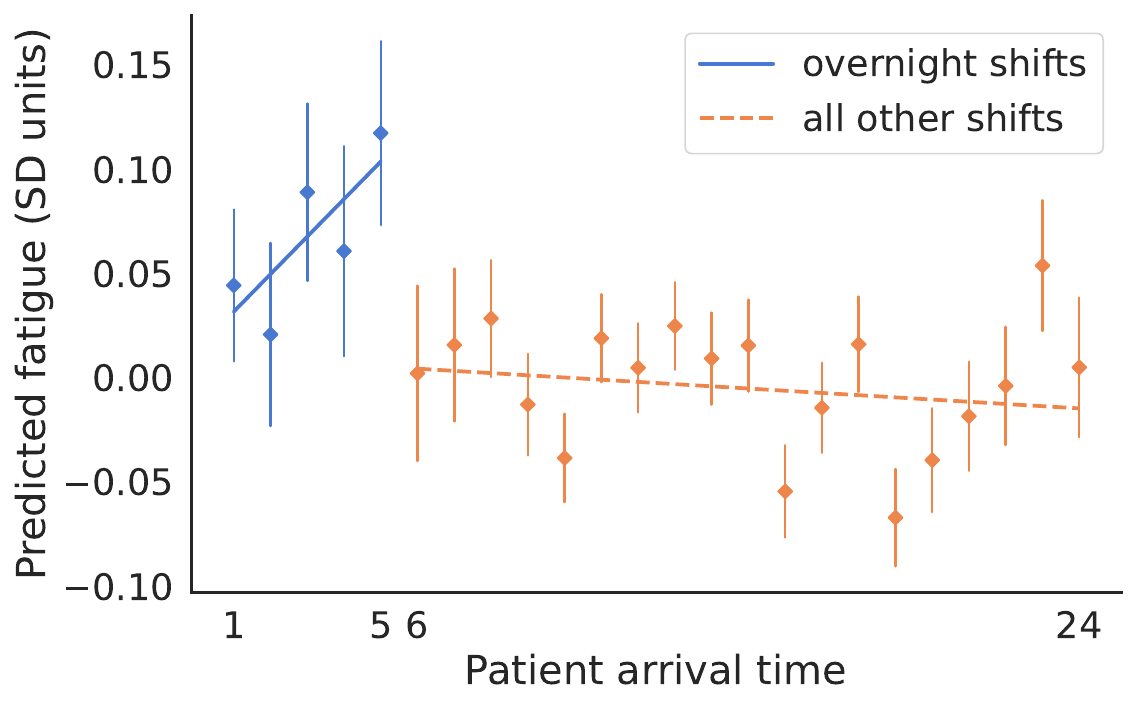}
    \caption{Predicted fatigue vs. patient arrival time. The $y$-axis shows model predictions, in standard deviation units. The $x$-axis shows patient arrival time. Patients arriving after 1am and before 6am (blue) can only be seen by the overnight physician, while other patients can be seen by any of the multiple physicians working at a given time (orange).    }
    \label{fig:next_hour_pred} 
\end{figure}

Figure~\ref{fig:next_hour_pred} shows model predictions graphically as a function of patient arrival time. Consistent with the regression results, model predictions are greater on average for patients arriving between 1am and 6am; in fact, predictions increase over that time period (slope: 1.79, std err: 1.34), vs stable model predictions over the rest of the day (slope: -0.10, std err: 0.12).

Our second measure of circadian disruption is the variance of physician shift start times over the prior week (including the current shift). The intuition is that a physician who has worked three prior shifts starting at 7am, and is now starting again at 7:00am, is less disrupted than a physician who worked prior three shifts starting at 7:00am, 5:00pm, and 11:00pm. Table~\ref{tab:result-table}, Column 3 
again shows a significant coefficient on model predictions. 

Finally, we link model predictions to increasing fatigue over the course of a shift, measured by how many patients a physician has already seen on that shift (Table~\ref{tab:result-table}, Column 4), again showing a large and significant coefficient. This is reassuring: while predicted prior workload could be picking up on spurious correlations or other aspects of physician behavior besides fatigue, the fact that it correlates with several other more specific settings where fatigue is common, makes it more likely that we are indeed measuring fatigue and not some other phenomenon.

\section{Relationship of Measured Fatigue to Physician Decision-making}\label{decisions}

A common dilemma in studies of physician fatigue is the inability to measure \textit{actual} physician fatigue. Instead, they measure physician exposure to \textit{drivers} of fatigue like workload or circadian factors. Of course, these two are correlated---but given that inputs are an imperfect measure of outputs (given a range of idiosyncratic shocks to fatigue, from sleep quality to diet and exercise), studies of inputs may suffer from attenuation bias and fail to find an effect. As a result, most studies to date have found effects of (drivers of) fatigue on physician behavior, but little impact on patient outcomes \citep{landrigan2004effect, barger2006impact, lockley2004effect, institute2009resident, ayas2006extended, arnedt2005neurobehavioral}. 

We hypothesized that our measure of fatigue might address some of these shortcomings. Our measure can be calculated at the patient-note level, and importantly---as noted in Section \ref{analytic} and Appendix \ref{sec:model}---is a more accurate measure of `revealed' physician fatigue than inputs like workload, circadian disruption, etc. 

To test this, we build on prior work by \citet{mullainathan2022diagnosing} examining the quality of a critical decision in the emergency setting: whether or not to test a patient for acute coronary syndromes (ACS, colloquially: heart attack). This decision is an important yet challenging one. ACS is important to diagnose in a timely manner, but testing for it is invasive and resource intensive. Research has shown widespread over-testing: some patients have such predictably low likelihood of testing positive that it’s really not worth doing the test, as measured by the yield of testing, i.e., the fraction of tests that come back positive \citep{hermann2013yield, stripe2013diagnostic}. Following this literature, we view yield as a patient-centered measure of the quality of physician decision-making: a higher yield means a higher rate of diagnosing ACS and unlocking enormous health benefits for the patient, while a lower yield means incurring the costs and risks of testing with no clear patient benefit.

Table~\ref{tab:workload-fatigue-precision} shows the results.
We first regress the yield of testing on the number of days worked in the prior week, controlling for time, patient, and physician effects. This simulates the standard way of testing the effect of physician fatigue on patient outcomes, with a coarse measure of workload that does not vary at the patient level. We find no effect: the coefficient on days worked is small and insignificant. We then repeat the analysis, substituting our note-based measure of fatigue, at the patient level, instead of the coarse measure of days worked. Here, the coefficient is large and significant: for each one standard deviation increase in fatigue, the yield of testing decreases by 17.9\%. This result indicates that fine-grained measures of fatigue like the one we use here are a promising new way to measure and elucidate the consequences of physician fatigue.

\begin{table*}[]
    \centering
    \begin{tabular}{@{}lcc@{}}
    \toprule
              & Effect of Workload     &     Effect of Predicted Fatigue     \\ 
              & on Yield of Testing &  on Yield of Testing \\\midrule
    Coefficient  & -0.0062     &  -0.2668$^{*}$ \\
        & (0.009)            & (0.130)        \\
              &                         \\
    \textbf{Controls}   &                         \\
    {} Time of day      & \multicolumn{1}{c}{YES}& \multicolumn{1}{c}{YES} \\
    {} Day of week      & \multicolumn{1}{c}{YES}& \multicolumn{1}{c}{YES} \\
    {} Week of year      & \multicolumn{1}{c}{YES}& \multicolumn{1}{c}{YES} \\
    {} Year      & \multicolumn{1}{c}{YES}& \multicolumn{1}{c}{YES} \\
    {} Demographics      & \multicolumn{1}{c}{YES}& \multicolumn{1}{c}{YES} \\
    {} Chief complaint       & \multicolumn{1}{c}{YES} & \multicolumn{1}{c}{YES}\\
    {} Physician & \multicolumn{1}{c}{YES}  & \multicolumn{1}{c}{YES}\\\bottomrule
    \end{tabular}
    \caption{Effect of workload and effect of predicted on yield of testing for heart attack (more precisely, the positive rate of tests done in the few days after the ED visit) ($n=1,017$). Column 1 shows a regression of testing yield on a physician's prior week workload, which is not significant. Column 2 shows a regression of testing yield on the predicted fatigue score, calculated on the basis of the patient's note, which is significant and negative. 
    }
    \label{tab:workload-fatigue-precision}
\end{table*}

\section{Correlation of Measured Fatigue with Patient Race and Other Factors}\label{race}

\begin{table*}[t]
\centering
\begin{tabular}{@{}lcc@{}}
\toprule
                         & Workload (Day)& Predicted Fatigue      \\ \midrule
Race (vs. White) &            &                \\
{} {} Black                  & -0.0009  & 0.0112***          \\
{} {}                   & (0.017)  & (0.001)              \\
{} {} Hispanic            & -0.0125     & 0.0192***          \\
{} {}                  & (0.018)   & (0.001)              \\
{} {} Other                & -0.0213    & 0.0086***          \\
{} {}                   & (0.027)   & (0.002)             \\
                         &            &                \\
Female (vs. Male)       & 0.0041  & -0.003**            \\
                 & (0.015) & (0.001)                \\
                         &            &                \\
Age                       & 0.0006 & -0.0003***        \\
                 & (0) & (0.00003)                  \\
                         &            &                \\
Intercept              & 3.2354***  & 0.569***         \\
                  & (0.084) & (0.005)               \\
                         &            &                \\
\textbf{Controls}         &            &                \\
{} Time of day           & YES & YES      \\
{} Day of Week           & YES & YES      \\
{} Week of Year           & YES & YES       \\
{} Year           & YES & YES       \\
{} Chief complaints                       & YES        & YES            \\
{} Physician                & YES        & YES            \\ \bottomrule
\end{tabular}
    \caption{Correlation of patient demographic factors with workload and predicted fatigue. Column 1 shows a regression of a physician's prior week workload on demographic factors, none of which are significant. Column 2 shows a regression of the predicted fatigue score, calculated on the basis of the patient's note, on demographics. Several factors are significant.}
\label{tab:dem-bias}
\end{table*}

Just as a note-based fatigue measure allows for fine-grained analyses of patient outcomes, it can also be used to explore how measured fatigue relates to patient characteristics, specifically demographics. We use electronic health record data on race and ethnicity to capture race, which is self-reported by patients within the categories provided by the health record. Given sample size considerations, we were unable to explore intersectional identities beyond Black, Hispanic, and White. Any patient who self-identified as Black or Hispanic was coded according to their self-identified group; patients who did not identify as any race or ethnicity other than White were coded as White for the purposes of the analysis. 

We first test whether actual physician workloads over the prior week are correlated with age, sex, race, and chief complaints, and find no relationship (Table~\ref{tab:dem-bias}, Column 1). Measured fatigue, on the other hand, is highly correlated with several of these factors (Table~\ref{tab:dem-bias}, Column 2). 

Relative to White patients, notes written about both Black and Hispanic patients were scored as having higher measured fatigue. To put these differences in context, we compare them to the average difference in measured fatigue for patients seen by the overnight physician, by regressing measured fatigue on race and an indicator for the overnight shift (Supplement Table~\ref{tab:race-tiredness-reg}, Column 3), with all the usual controls. We find that measured fatigue in notes written about Hispanic patients is nearly 4x as large as the overnight effect, while for Black patients it is over 2x. Differences in treatment of Black patients are by now well established; our results add to this literature by stressing that healthcare disparities for Hispanic patients are also a matter for concern \citep{flores2020disparities}. Importantly, this does not appear to be a language effect alone: we have access to a patient’s self-reported primary language, and the effect among Hispanic patients whose primary language is English vs. Spanish is similar (Supplement Table~\ref{tab:tiredness-race-language}).

\section{Features of Notes that Correlate with Fatigue}

To provide some intuition on the features of notes used by our model to predict fatigue, we show correlations of model features with physician workload (specified as an indicator of high- vs. low-workload, the label used for training the model) in Table~\ref{tab:corraltion-main}. 

The most highly correlated feature is the predictability of a note, as measured by note `perplexity.' Intuitively, this feature measures how hard it is for a large language model to predict a given word in a note, conditional on the preceding words in the note.\footnote{The model is fine-tuned in our data and uses the 1024 previous tokens to make the prediction; more details are in the Supplementary Material. Perplexity has mostly been used as a measure of the quality of large language models, a question we return to below.} We find that perplexity decreases with workload, implying that fatigued physicians tend to write more predictable notes. This could suggest that tired physicians capture less information: if the words a physician writes about one patient can be predicted based on a corpus of clinical notes about other patients, the physician may be viewing the patient in a generic rather than a personalized way.\footnote{Related work using text analysis to differentiate between written descriptions of real vs. imagined events has also found strong correlations with note perplexity ~\citep{sap2020recollection}.} Related to predictability, the readability of notes, based on the complexity of sentences and words, is lower when physicians are fatigued.

\begin{table*}[]
\centering
\begin{tabular}{@{}lc@{}}
\toprule
Feature                                    & Correlation with  \\ 
                                    & physician workload  \\ \midrule
\textbf{Word Un-predictability}                    &                    \\
{} Perplexity (log)                             & -0.092$^{***}$ \\
                                           &                  \\
\textbf{Cognitive words (LIWC)}                   &                     \\
{} Insight (fraction)               & -0.09$^{***}$  \\
{} Certainty (fraction)            & 0.075$^{***}$  \\
                                           &                  \\
\textbf{Pronouns (LIWC)}                      &                   \\
{} First person singular pronouns (fraction) & -0.07$^{***}$   \\
{} Impersonal pronouns (fraction)           & -0.049$^{***}$  \\
                                           &                  \\
\textbf{Affect (LIWC)}                       &              \\
{} Anger (fraction)                & 0.025$^{***}$ \\
\textbf{}                                  &               \\
\textbf{Readability}                       &         \\
{} Flesch-Kincaid grade                       & -0.05$^{***}$  \\ \bottomrule
\end{tabular}
\caption{Features used by the model: Pearson's correlation with physician fatigue, as measured by workload over the prior week. More details are in the Supplementary Materials.}
\label{tab:corraltion-main}
\end{table*}

We also note the fraction of words related to insight and cognition, as categorized by the standard LIWC corpus \citep{pennebaker2001linguistic}, are linked to model predictions. The insight category includes words such as “believe,” “reveal,” “think,” and “explain,” and the certain category includes words such as “clear,” “certain,” “apparent,” and “never.” Fatigued physicians tend to use fewer words in the insight category, and more words in the certain category. Related, we also find that fatigued physicians are less likely to use first person singular pronouns: many insight words require association with first person singular pronouns such as ``I believe.'' Other work has linked similar patterns with withdrawal and less personal agency~\citep{konopasky2020linguistic}. Finally, we find that fraction of words related to anger, for example lying, assault, and threat, are also positively correlated with fatigue.\footnote{Recall that we find no correlation between patients' reason for visiting the ED (their chief complaint) and physician workload, suggesting that differential assignment of assault cases to physicians does not explain this finding.}

\section{Implications for LLMs}

Our results have important implications for the emerging literature on LLMs trained on medical notes~\citep{jiang2023health}. First, while LLMs excel at many tasks related to language, we show that they are substantially worse than a supervised model at distinguishing notes written by high vs. low workload physicians. Using Vicuna-7B \citep{vicuna2023} for zero-shot classification, the AUC-ROC was significantly lower than our approach (53.5\%;, 95\% CI: 53.3-53.9; see the Supplementary Materials \ref{vicuna}). This is perhaps unsurprising as fatigue is a latent, rather than an explicit, component of physician notes and there are few training data to learn from; it also illustrates that specialized models still have many advantages for medical tasks, as has been noted in~\citet{lehman2023we}. 

Second, as shown above, a key feature linked to fatigue in our model is the predictability of the next word in a note. Text generated by LLMs is, by construction, predictable: next-word prediction is the basis for the core training task around which LLMs are built. This raises the possibility that LLM-generated notes may be characterized by many of the same features that make for fatigued-appearing physician notes. To test this, we prompt an LLM with the first sentence of a real note, and ask it to complete the patient history. We then compare this to the real patient history and quantify measured fatigue for each pair. Patient histories generated by LLMs have 74\% higher predicted fatigued than genuine physician notes, and in particular have higher scores on predictability and fraction of anger words. This illustrates some of the dangers of relying on LLMs as a substitute for physicians in generating clinical notes: to the extent we believe that fatigue generally leads to worse notes, LLMs may be producing lower-quality information than is initially apparent.

Furthermore, as writing is an important form of thinking~\citep{menary2007writing}, automation may cause physicians to skip some of the active thinking components of the patient encounter.
This echoes recent work showing that LLMs can perpetuate racial biases in generated medical texts \citep{adam2022write, zack2023coding}. It also calls into question the strategy of relying on humans to evaluate the quality of text produced by LLMs, which is pervasive in the current literature~\citep{nayak2023comparison,singhal2023large}. While we were not able to explore this in the current work, we suspect clinicians may be just as unable to identify fatigued notes as LLMs are. And given the possibility that fatigued notes miss important details --- as suggested by the correlation between note-based fatigue and low-yield testing for heart attack --- new benchmarks are urgently needed to assure quality before LLMs become widely deployed. 

Last but not least, instead of the straightforward yet potentially harmful application of automating note-writing~\citep{zack2023coding}, LLMs provide opportunities to transform the information solicitation process in physician-patient encounters. The predictable information in the notes may not require precious physician-patient interaction to acquire, while LLMs may help physicians identify valuable information to solicit by suggesting questions that lead to information with low predictability. While rested physicians may write notes that are less predictable, which may correlate with lower readability, LLMs may provide effective ways to separate the process of meeting patients and writing notes from the process of reading notes by simplifying the language when presented to patients or other physicians.

In short, we advocate responsible adoption of LLMs as a writing assistance tool in the healthcare domain. It is important to ensure that LLMs serve as an augmentation tool to improve the note-taking practice, without eroding the agency of physicians and reducing the value of information in clinical notes.

\section{Conclusion}

We demonstrate the potential of detecting physician fatigue using the notes that they wrote.
Our predicted fatigue allows us to reveal connections with decision quality and disparity between races. Our finding highlights the role of clinical notes not only as a medium of storing medical information but also a window into physician decision making.

\bibliography{refs}%
\clearpage
\appendix
\section{Behavioral Model for the Prediction Task}
\label{sec:model}

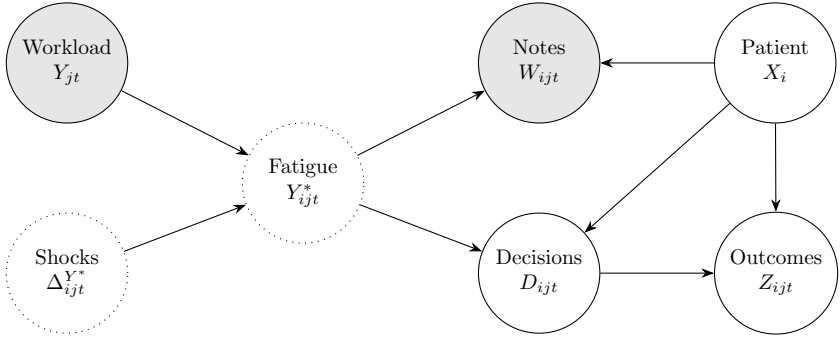
\begin{figure}
    \centering
    \begin{tikzpicture}[>=Stealth, node distance=1.5cm, scale=0.8, every node/.style={scale=0.8}]
    
        \node (Ystar) [circle, draw, minimum size=2cm, dotted, align=center] {Fatigue \\ \( Y^*_{ijt} \)};
        \node [left=of Ystar, yshift=2cm, circle, draw, minimum size=2cm, fill=gray!20, align=center] (Yjt) {Workload \\ \( Y_{jt} \)};
        \node [left=of Ystar, yshift=-1.5cm, dotted, circle, draw, minimum size=2cm, align=center] (DeltaYijt) {Shocks \\ \( \Delta^{Y^*}_{ijt} \)};
        \node [right=of Ystar, yshift=2cm, circle, draw, minimum size=2cm, fill=gray!20, align=center] (Wijt) {Notes \\ \( W_{ijt} \)};
        \node [right=of Ystar, yshift=-1.5cm, circle, draw, minimum size=2cm, align=center] (Dijt) {Decisions \\ \( D_{ijt} \)};
        \node [right=of Wijt, circle, draw, minimum size=2cm, align=center] (Xi) {Patient \\ \( X_{i} \)};
        \node [right=of Dijt, circle, draw, minimum size=2cm, align=center] (Zijt) {Outcomes \\ \( Z_{ijt} \)};
        
        \draw[->] (Yjt) -- (Ystar);
        \draw[->] (DeltaYijt) -- (Ystar);
        \draw[->] (Ystar) -- (Wijt);
        \draw[->] (Ystar) -- (Dijt);
        \draw[->] (Xi) -- (Dijt);
        \draw[->] (Xi) -- (Wijt);
        \draw[->] (Xi) -- (Zijt);
        \draw[->] (Dijt) -- (Zijt);

    \end{tikzpicture}
    \caption{Variable causal relationships. Dotted nodes are unobserved, shaded nodes are components of our statistical model $Y_{jt} = g(W_{ijt})$.}
    \label{fig:dag}
\end{figure}

Here we provide a more detailed argument for why we believe our predictions will be a \textit{better} measure of a physician's true fatigue when writing a given note than prior workload. 

Recall that doctor $j$ sees patient $i$ with characteristics $X_i$ during shift $t$. We assume past-week workload, denoted as $Y$, affects ``true'' fatigue $Y^*$,
$$ Y^*_{ijt} = Y_{jt} + \Delta^{Y^*}_{ijt} $$
where $\Delta^{Y^*}$ is a set of idiosyncratic random shocks and coping strategies affecting true fatigue when the physician sees patient $i$. We assume (and verify) $Y \perp X$. We also assume that $Y_{jt}$ affects the written note $W_{ijt}$ and physician decision quality $D_{ijt}$ only via $Y^*$. Decisions $D_{ijt}$ and patient factors $X_i$ jointly determine patient outcomes $Z_{ijt}$ at that visit. These relationships are summarized in Supplement Figure \ref{fig:dag}. 

Intuitively, because our statistical model takes the form  $Y_{jt} = g(W_{ijt})$
and the label $Y_{jt}$ is the same for all patients $i$ seen by doctor $j$ on her shift $t$, it would seem that following a ``perfect'' model would predict a constant $\hat{Y}_{ijt} = E[\hat{Y}_{ijt}]$. However, note that both $Y_{jt}$ and $\Delta^{Y^*}_{ijt}$ affect $Y^*_{ijt}$, and there are no other paths by which either can affect notes except via $Y^*_{ijt}$. As a result, $g(\cdot)$ learns about features of notes $W$ that predict workload $Y$; but because this relationship is mediated through $Y^*$, it also learns about the idiosyncratic shocks $\Delta^{Y^*}$ that affect $W$ through the same channel $Y^*$ (and it does so even though $Y \perp \Delta^{Y^*}$, by construction). 

This has a key implication for a model that uses $W_{ijt}$ to predict $Y_{jt}$: it will \textit{not} predict $Y_{jt}$ accurately because $W_{ijt}$ is correlated with $\Delta^{Y^*}_{ijt}$, while $Y_{jt}$ is not. Notes can be written
$$ W_{ijt} = Y^*_{ijt} + X_i = (Y_{jt} + \Delta^{Y^*}_{ijt}) + X_i$$
which means that predictions 
$$\hat{Y}_{ijt} \approx E[Y_{ijt} \mid W_{ijt}] = E[Y_{ijt} \mid Y_{jt}, \Delta^{Y^*}_{ijt}, X_i] = E[Y_{ijt} \mid Y_{jt}, \Delta^{Y^*}_{ijt}]$$ 
where $X_i$ is dropped because (as we show) $Y \perp X$. As a result, $\hat{Y}_{ijt}$ will contain ``errors'' when evaluated against the original label $Y_{jt}$, because of idiosyncratic variations in true fatigue $ \Delta^{Y^*}_{ijt}$. Indeed, as the influence of $\Delta^{Y^*}_{ijt}$ on $Y^*_{ijt}$ grows large relative to $Y_{jt}$, our predictions will approach closer to $\hat{Y}^*_{ijt}$, but contain larger ``errors'' relative to the ``true'' label $Y_{jt}$.

A formal proof of the statement in a simple uni-dimensional linear setting follows. Assume $W_{ijt} = (Y_{jt} + \Delta^{Y^*}_{ijt}) A$, then the regression coefficients of $Y \sim W$ is (we omit the subscripts for ease of notation): 

\begin{align*}
    \hat{\beta} &= (W^TW)^{-1}W^TY \\
    & = (A^T(Y+\Delta^{Y^*})^T(Y+\Delta^{Y^*})A)^{-1} (A^T(Y+\Delta^{Y^*})^T) Y \\
    & = \frac{1}{A} \cdot \frac{1}{Y^TY + (\Delta^{Y^*})^T \Delta^{Y^*}} \cdot (Y^TY).
\end{align*}
Recall $A$ is a scalar in this case. It follows
\begin{align*}
    \hat{Y} &= W\hat{\beta} \\
    & = (Y+\Delta^{Y^*})A \cdot \frac{1}{A} \cdot \frac{1}{Y^TY + (\Delta^{Y^*})^T \Delta^{Y^*}} \cdot (Y^TY) \\
    & = \frac{Y^TY}{Y^TY + (\Delta^{Y^*})^T \Delta^{Y^*}} (Y+\Delta^{Y^*}) 
\end{align*}
We observe that $\hat{Y}$ is a shrunk version of $Y+\Delta^{Y^*}$, and how well it captures $Y$ is determined by the relative magnitude of $Y$ and $\Delta^{Y^*}$.

As a result, our model predictions could be ``wrong'' (with respect to $Y_{jt}$) because they are good---i.e., closer to $Y^*_{ijt}$. Alternatively, the model could be wrong in less useful ways---it could simply be making bad predictions. Importantly, the graph above provides an empirical test of which is more true: good predictions will correlate with patient outcomes $Z_{ijt}$, via $Y^*_{ijt}$, which we test in Section \ref{decisions}.

Lastly, this model helps us contextualize the fact that patient characteristics $X_i$ may correlate with $\hat{Y}_{ijt}$ but not $Y_{jt}$. In our model, this can only be because of correlations between $X_i$ and the idiosyncratic shocks to fatigue $\Delta^{Y^*}_{ijt}$ that can vary at the patient level and are incorporated into $\hat{Y}_{ijt}$. We test 
that these characteristics are not correlated with workload. Failing this, it is possible---as considerable research has demonstrated both inside and outside of the hospital---that latent psychological factors are present at the patient level that mimic fatigue when dealing with demographically different patients.

\section{Dataset}

Our physician notes dataset comes from a high-ranked hospital in an urban area of the United States from 2010 to 2012. See \tableref{tab:data-summary-split} for details of the dataset split and the basic demographics statistics.
We show 10 most frequent chief complaints in \tableref{tab:top-10-cc}.
The notes collected are written in the same day of the patient encounter except 32 notes (0.02\%) that are written after the encounter date.

\begin{figure}
    \centering
    \includegraphics[width=1\textwidth,trim=2cm 1.8cm 2cm 1cm,clip]{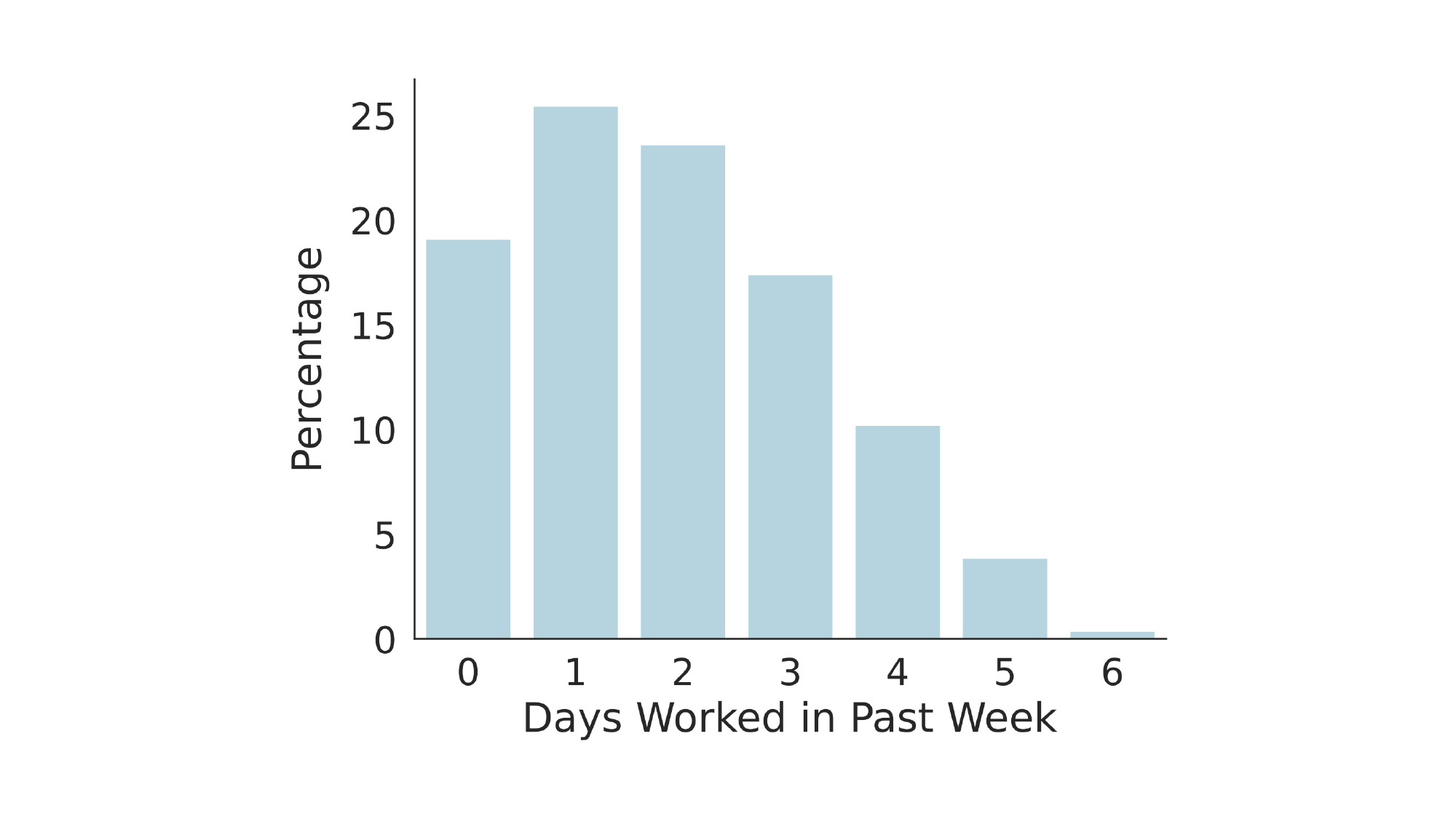}
    \caption{Histogram of days worked in the seven days prior to writing a note (including the day when the patient encounter happened; 0 means that the patient encounter happened on the first day that the physician worked in the seven-day window
    ). 
    }
    \label{fig:histogram}
\end{figure}

\paragraph{Note Format} 
Generally, physician notes are semi-structured: they are grouped into sections and each section includes free texts. 
The most common sections are \texttt{History of Present Illness}, 
\texttt{Physical Examination}, and \texttt{Assessment and Plan}.
First, the \texttt{History of Present Illness} section describes the reason for the encounter.
Then, 
\texttt{Physical Examination} section is also included to understand what preliminary tests have been done in this encounter. 
Finally, the \texttt{Assessment and Plan} section presents the thought processes for the final diagnosis.

\begin{table*}[h]
    \centering
    \begin{tabular}{lr}
        \toprule
                       Chief Complaint &   count \\
        \midrule
                           abd pain & 35895 \\
              chest or esophag pain & 22679 \\
                            dyspnea & 16381 \\
                          back pain & 12920 \\
                       fever chills & 11389 \\
                               fall & 10382 \\
        elbow wri hand finger cmplt & 10266 \\
                          leg cmplt &  9192 \\
               foot ankle toe cmplt &  9091 \\
                           headache &  8691 \\
        \bottomrule
        \end{tabular}
        
    \caption{Top 10 frequent chief complaints.}
    \label{tab:top-10-cc}
\end{table*}

\begin{figure}
  \centering
  \includegraphics[width=1\columnwidth]{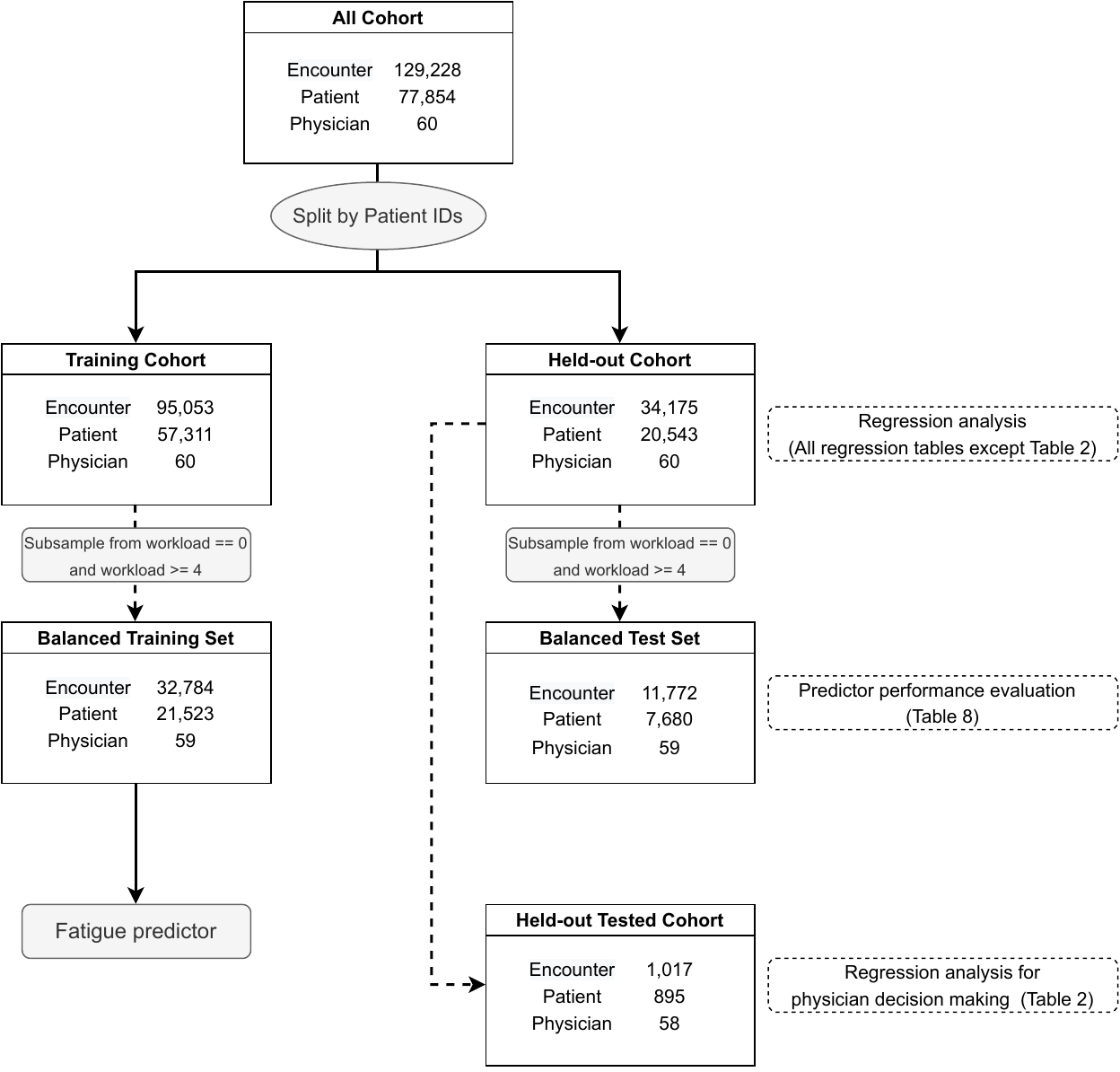}
  \caption{Data diagram. The dataset contains 129,228 encounters from 77,854 patients and 60 physicians.
  }
  \label{fig:data-diagram}
\end{figure}

\paragraph{Data Split} Fig.~\ref{fig:data-diagram}  illustrates the data splitting workflow in this work.

\begin{table*}[htbp]
  \centering
  \label{tab:data-summary}
  \begin{tabular}{lrrr}
    \toprule
    \textbf{Category} & \textbf{All} & \textbf{Training set} & \textbf{Heldout set}\\
    \midrule
    \textbf{Encounters} & 129,228 & 95,053 & 34,175\\
    \textbf{Patients} & 77,854 & 57,311 & 20,543\\
    \textbf{Physicians} & 60 & 60 & 60\\
    \midrule
    \textbf{Demographics} & &\\\\
    Age & 41.8 & 41.9 & 41.8\\
    White & 0.50 & 0.50 & 0.50\\
    Black & 0.22 & 0.22 & 0.21\\
    Hispanic & 0.18 & 0.18& 0.18\\
    Female & 0.60 & 0.60 & 0.60\\
    Male & 0.40 & 0.40 & 0.40\\
    \bottomrule
  \end{tabular}
  \caption{Summary of basic statistics.}
  \label{tab:data-summary-split}
\end{table*}

\section{Fatigue Predictor}
\label{sec:predictor}

\subsection{Features}
For every note used in this paper for the tiredness predictor, we drop the beginning and the end of note which are generated by the system.
Instead of directly using textual content in the physician note, we select features of note contents to avoid spurious correlation that can be inferred from the actual text and improve the interpretability of the predictor.
We include basic statistics of notes, i.e., note length, fraction of stopwords, fraction of medical words, and readability.
Also, we use LIWC \citep{pennebaker01} lexicon to capture cognitive and affective processes of physiciains.
Finally, we introduce a novel feature of note predictability obtained from a fine-tuned large language model, GPT-2 with 117 million parameters \citep{radford2019language} (see more details later).
See Table~\ref{tab:feature-value} for feature values.

After extracting the features, we standardize the features to have zero mean and unit variance. We also include the dummy chief complaint features to control for the effect of chief complaints on predicting the fatigue of notes.
Table~\ref{tab:feature-value} shows basic statistics for different features.

\begin{table*}
\centering
\begin{tabular}{lrr}
    \toprule
    {} &      mean &     standard error \\
    \midrule
    \textbf{Note characteristics} & \textbf{}& \textbf{}\\
    {} log perplexity                   &    1.1969 &  0.0008 \\
    {} note length                   &  328.0966 &  0.4137 \\
    {} fraction of stopwords              &    0.4027 &  0.0002 \\
    {} fraction of medical words           &    0.4815 &  0.0001 \\
    \textbf{LIWC Pronoun} & \textbf{}& \textbf{}\\
    {} fraction of pronoun                 &    0.0755 &  0.0001 \\
    {} {} fraction of first person singular pronouns   &    0.0030 &  0.0000 \\
    {} {} fraction of first person plural pronouns   &    0.0024 &  0.0000 \\
    {} {} fraction of second person pronouns &    0.0000 &  0.0000 \\
    {} {} fraction of third person singular pronouns   &    0.0463 &  0.0001 \\
    {} {} fraction of third person plural pronouns &    0.0005 &  0.0000 \\
    {} {} fraction of impersonal pronouns  &    0.0229 &  0.0000 \\
    \textbf{LIWC Affect} & \textbf{}& \textbf{}\\
    {} fraction of affective processes   &    0.0360 &  0.0000 \\
    {} {} fraction of positive emotions   &    0.0151 &  0.0000 \\
    {} {} fraction of negative emotions   &    0.0204 &  0.0000 \\
    {} {} fraction of anxiety &    0.0055 &  0.0000 \\
    {} {} fraction of anger  &    0.0009 &  0.0000 \\
    {} {} fraction of sadness &    0.0029 &  0.0000 \\
    \textbf{LIWC Cognitive} & \textbf{}& \textbf{}\\
    {} fraction of cognitive processes &    0.1260 &  0.0001 \\
    {} {} fraction of insight &    0.0154 &  0.0000 \\
    {} {} fraction of causation   &    0.0049 &  0.0000 \\
    {} {} fraction of discrepancy   &    0.0064 &  0.0000 \\
    {} {} fraction of tentative  &    0.0221 &  0.0000 \\
    {} {} fraction of certainty  &    0.0047 &  0.0000 \\
    {} {} fraction of inhibition  &    0.0056 &  0.0000 \\
    {} {} fraction of inclusive &    0.0581 &  0.0000 \\
    {} {} fraction of exclusive   &    0.0196 &  0.0000 \\
    \textbf{Readability} & \textbf{}& \textbf{}\\
    {} Flesch-Kincaid grade &    8.1098 &  0.0037 \\
    \bottomrule
\end{tabular}
\caption{Feature statistics for the tiredness predictor over the whole dataset. 
}
\label{tab:feature-value}
\end{table*}

\paragraph{Perplexity and language modeling details}
A language model predicts the probability of the next token, 
and thus assigns probability to a sentence or document:
$p(X) = \prod_{i=1}^{n} p\left(x_{n} \mid x_{1}, \ldots, x_{n-1}\right)$,
where $X$ is the document and $x_i$ denotes $i$-th token in the document.
Perplexity is then computed as
$\texttt{perplexity}(X) 
= 2^{-\frac{1}{N}\texttt{log}_2 p(X)}$.
A lower perplexity indicates that the text is more predictable by the language model.

Recent studies in large-scale language models show that finetuning from the pretrained checkpoint can improve the domain-specific language modeling \citep{Gururangan2020DontSP,Wang2021DomainSpecificPF}.
In this work, 
we finetune GPT-2 (117M)
on our training set and compute the perplexity of each note (document) in the whole dataset.
The greater the log perplexity, the lower the {\em note predictability}.

\subsection{Classification Experiment Setup}\label{vicuna}
The balanced predcitor dataset is derived from the original dataset in \tableref{tab:data-summary-split}. 
We define a note as fatigued note if the note is written by a physician working last 4 days before today. In contrast, we define a note as non-fatigued note if the note is written by a physician working 0 days before today in a 7-day span.
For each chief complaint, we sample the same number of fatigued and non-fatigued notes from the original training and hold-out set to form the balanced dataset.
Given the fatigue dataset, we use the logistic regression model with regularization as our classifier. 
We conduct hyperparameter search on the regularization term with 5-fold cross validation on the training set.
\paragraph{Baseline Model}
For the baseline model, we only use chief complaint as the input feature.
We use the same logistic regression model with regularization as our classifier.

\paragraph{Prompt for the Vicuna-7B model.}
To see how pretrained language models can be used for the fatigue prediction task in a zero-shot manner, we use the Vicuna-7B model to perform the task.
We sample 1000 pairs of fatigued and non-fatigued notes from the balanced test set and ask the Vicuna-7B model to predict which note is written by a more fatigued physician with the following prompt.

\scriptsize
\begin{verbatim}
Prompt:
Note 1: {shuffled_notes[0]} 


Note 2: {shuffled_notes[1]} 


Task:
Analyze the above two physician notes and assess 
which one appears to be written by a more fatigued physician. 
Answer the question at the end by selecting either [Note 1] or [Note 2].
Only reply the answer.
Do not include any other information.
\end{verbatim}
\normalsize

\subsection{Model Performance}
Table~\ref{tab:performance} shows detailed performance for the prediction task on the balanced held-out test set. Our model with note features outperforms the baseline model with only chief complaint as the input feature.

\subsection{Model Coefficients}
\figref{tab:coef} shows the logistic regression model coefficients excluding chief complaints. Note predictability has the highest coefficient of -0.271, the same as the feature correlation analysis. Note that the input features are standardized. 

\begin{table}[]
\centering
\begin{tabular}{@{}lc@{}}
\toprule
\textbf{Feature}                           & \textbf{Coefficient} \\ \midrule
\textbf{Note Feature}                      &                      \\
{} note length                                & -0.04971             \\
{} log perplexity                             & -0.271               \\
{} fraction of stopwords                      & 0.12594              \\
{} fraction of medical words                  & 0.04612              \\
                                           &                      \\
\textbf{LIWC Pronoun}                      &                      \\
{} fraction of pronoun                        & 0.06129              \\
{} {} fraction of first person singular pronouns & -0.14271             \\
{} {} fraction of first person plural pronouns   & 0.06668              \\
{} {} fraction of second person pronouns         & -0.01114             \\
{} {} fraction of third person singular pronouns & -0.07235             \\
{} {} fraction of third person plural pronouns   & 0.04096              \\
{} {} fraction of impersonal pronouns            & -0.10904             \\
                                           &                      \\
\textbf{LIWC Affect}                       &                      \\
{} fraction of affective processes            & 0.02486              \\
{} {} fraction of positive emotions              & -0.05055             \\
{} {} fraction of negative emotions              & -0.08713             \\
{} {} fraction of anxiety                        & 0.00322              \\
{} {} fraction of anger                          & 0.07171              \\
{} {} fraction of sadness                        & 0.03816              \\
                                           &                      \\
\textbf{LIWC Congnitive}                   &                      \\
{} fraction of cognitive processes            & 0.01288              \\
{} {} fraction of insight                        & -0.12494             \\
{} {} fraction of causation                      & 0.07551              \\
{} {} fraction of discrepancy                    & 0.05359              \\
{} {} fraction of tentative                      & 0.01638              \\
{} {} fraction of certainty                      & 0.15804              \\
{} {} fraction of inhibition                     & 0.04712              \\
{} {} fraction of inclusive                      & -0.02781             \\
{} {} fraction of exclusive                      & 0.0368               \\
                                           &                      \\
\textbf{Readability}                       &                      \\
{} Flesch Kincaid grade                       & -0.0112              \\ \bottomrule
\end{tabular}
\caption{Fatigue model feature coefficients.}
\label{tab:coef}
\end{table}

\begin{table*}[]
    \centering
    \begin{tabular}{@{}lccc@{}}
    \toprule
             & CC Baseline & CC + Note Features & Vicuna-7B Zero-shot*\\ \midrule
    AUC-ROC  & 50.0             & 60.1             &  53.5              \\
    Accuracy & 50.1             & 57.2             &      53.2          \\ 
    F1 & 38.9             & 56.6                &    49.1         \\ \bottomrule
    \end{tabular}
    \caption{Model performance on fatigue prediction. Chief Complaint (CC) baseline is the chance level. The pretrained instruct large language model Vicuna-7B achieves better than random accuracy and AUC-ROC scores in the zero-shot manner. The proposed fatigue predictor with note features outperforms both baseline model and the large language model. 
    * Note that Vicuna-7B is tested on 1000 pairs of fatigue and relaxed notes sampled from the balanced test set. 
     }
    \label{tab:performance}
\end{table*}

\section{Regression Details}

We present details on regression analysis presented in the group.

\paragraph{Control variables}
\begin{enumerate}
    \item Time control includes time of day, day of week, week of year, and year as categorical variables.
    \item Chief complaint control includes binarized chief complaints. An encounter can have multiple chief complaints.
    \item Physician control uses the physician id as a categorical variable.
    \item Patient demographics control uses the patient sex and race as categorical variables and patient age as the numerical variable.
\end{enumerate}

Due to the small number of tested observations in the test set for the regression on the yield of testing ($n=1017$), we only control for chief complaints and physicians with at least 30 occurrences.

\paragraph{Circadian disruption: Shift start time variance definition}
We compute the variance of starting time (hour of day) of the shifts in the past week including the current shift as one of our circadian disruption measurements. To mitigate the effect of midnight time change leads (23 vs. 0) to a huge variance, we shift the starting time back by 6 hours. For instance, two night shifts originally beginning at 23:00 and 00:00 are 17 and 18 in the adjusted schedule for the sake of computing the variance. Empirically, very few shifts started between 0 and 6am.

\paragraph{Dataset}
Regressions are done with our test set with 34,175 observations, except the yield of testing regressions, which only includes 1,017 observations of the tested cohort in the test set, and sanity check regression in \tableref{tab:sanity-workload} are done with high- and low-workload cohorts ($n=43,730$).

\section{Robustness Checks for the Tiredness Predictor}
\label{sec:sanity_check}

\paragraph{Workload vs. patient demographics and chief complaints}
We regress patient demographics on our target label, i.e., workload, to ensure no correlation between workload and patient demographics.
Table~\ref{tab:sanity-workload} that workload is not correlated with patient demographics.

\begin{figure}
    \centering
    \includegraphics[width=.8\columnwidth]{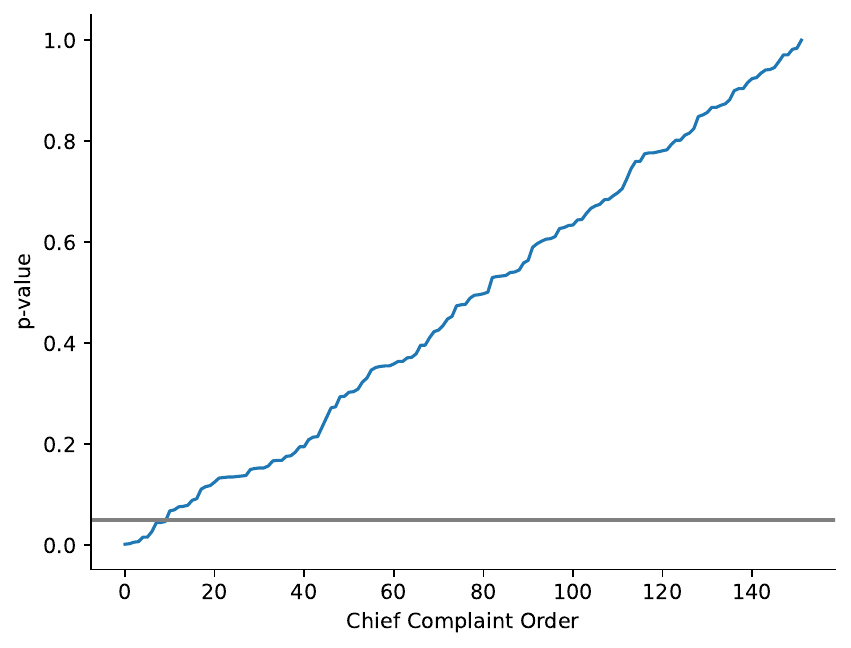}
    \caption{Sanity check for chief complaints. In only 10 out of 154 chief complaints, the workload coefficients have a significant effect, which present 6.5\% of all chief complaint count in the whole dataset, which is below 5\% out of randomness with $p$=0.05.}
    \label{fig:sanity-cc}
\end{figure}

We then regress chief complaints on workload and show that the correlation with chief complaints is not significant (\figref{fig:sanity-cc}).

\begin{table}[]
\centering
\begin{tabular}{@{}lccc@{}}
\toprule
                                    & Is Female            & Is White             & Age                  \\ \midrule
Workload                            & -0.0055              & -0.0023            & -0.1562              \\
                                    & (0.006)              & (0.006)              & (0.189)              \\\\
Intercept                           & 0.5694***            & 0.4724***            & 40.333***            \\
                                    & (0.029)              & (0.03)              & (0.961)              \\\\
\textbf{Controls}                             & \multicolumn{1}{l}{} & \multicolumn{1}{l}{} & \multicolumn{1}{l}{} \\
{} Time of Day & YES                  & YES                  & YES                  \\
{} Day of Week                         & YES                  & YES                  & YES                  \\
{} Week of Year                        & YES                  & YES                  & YES                  \\
{} Year                                & YES                  & YES                  & YES                  \\
{} Chief Complaint                     & YES                  & YES                  & YES                  \\
{} Physician                           & YES                  & YES                  & YES                  \\ \bottomrule
\end{tabular}
\caption{Sanity checks of regressing patient demographics on the high- and low-workload indicator on all cohorts (n=43,730). The regression results show that the target label (workload indicator) of our predictive model is not correlated with patient demographics.}
\label{tab:sanity-workload}
\end{table}

\paragraph{Variance for patient arriving at the same time}

\begin{figure}
    \centering
    \includegraphics[width=.8\columnwidth]{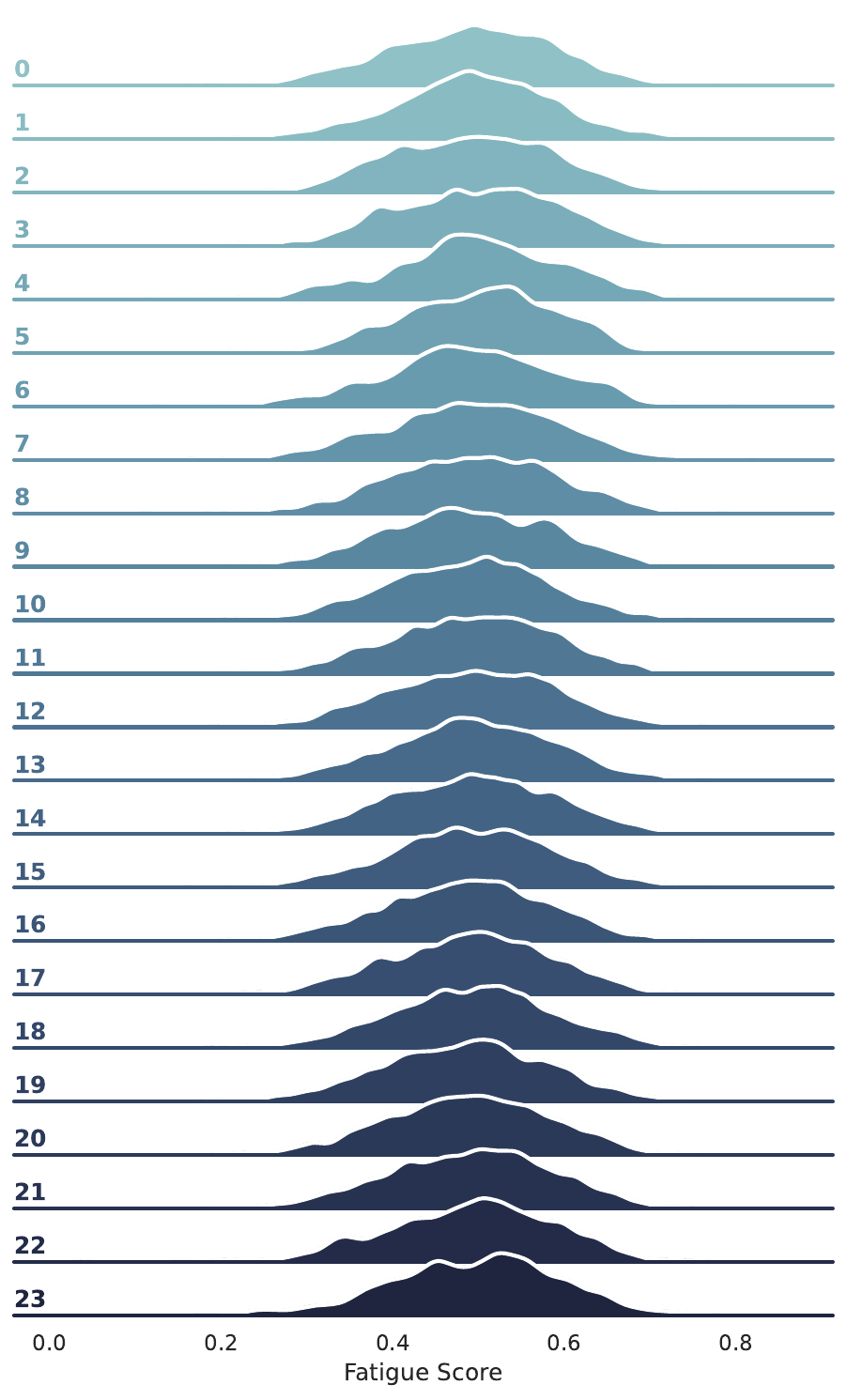}
    \caption{Fatigue score distribution over time of day.}
    \label{fig:time-fatigue-dist}
\end{figure}

Fig.~\ref{fig:time-fatigue-dist} shows that for each patient arriving time
there exists substance variance between estimated fatigue from notes. 

\section{Additional Analyses on Racial Biases}

\paragraph{Contextulizing the effect of racial bias}

Table~\ref{tab:race-tiredness-reg} contextualizes the effect of racial bias against that of the overnight indicator. Note that Fig.~\ref{fig:next_hour_pred} shows that the overnight indicator only captures the average effect over the entire night.

\begin{table*}[]
    \centering
    \small
    \begin{tabular}{@{}lccc@{}}
        \toprule
                       & Workload (Day) & Predicted Fatigue & Predicted Fatigue             \\ \midrule
        Race (vs. White) && & \\
        {} Black          &-0.0021 & 0.0118*** & 0.0119***             \\
                 & (0.017) & (0.001)     & (0.001)                 \\\\
        {} Hispanic       &-0.0144 & 0.0202*** & 0.0202***             \\
                 &(0.018) & (0.001)     & (0.001)                 \\\\
        {} Other          &-0.0228 & 0.0094*** & 0.0094***             \\
                 & (0.027) & (0.002)     & (0.002)                 \\\\
                       & &           &                       \\
        Is Night Shift  &  -    &  -         & 0.0052**             \\
                  & - &   -        & (0.002)                 \\
                        &&           &                       \\
        Intercept       &3.2633***& 0.5535*** & 0.5531***              \\
                  &(0.081)& (0.005)     & (0.004)                 \\
        &&           &                       \\
        \textbf{Controls}      &  &           &                       \\
        
        {} Time of day           & Yes & Yes       & No \\
        {} Day of Week           & Yes & Yes       & Yes  \\
        {} Week of Year           & Yes & Yes       & Yes  \\
        {} Year           & Yes & Yes       & Yes  \\
        {} Chief complaint & Yes& Yes       & Yes                   \\
        {} Physician       & Yes& Yes       & Yes                   \\  \bottomrule
    \end{tabular}
    \caption{Racial bias on predicted fatigue. We omitted gender in this set of regressions.}
    \label{tab:race-tiredness-reg}
\end{table*}

\paragraph{Checking the effect of language}
The racial biases in the predicted fatigue are similar between Hispanic patients with different self-reported use of languages (Hispanic and English). See \tableref{tab:tiredness-race-language}.
\begin{table*}[]
    \centering
    \begin{tabular}{@{}lc@{}}
    \toprule
                       & Predicted Fatigue \\ \midrule
    Race (vs. White) & \\
    {} Black              & 0.0119*** \\
                 & (0.001)     \\
   {}  Hispanic - English & 0.0192*** \\
                 & (0.001)     \\
    {} Hispanic - Spanish & 0.0211*** \\
                 & (0.002)     \\
    {} Other              & 0.0117*** \\
                 & (0.002)     \\
                       &           \\
    Intercept          & 0.5536*** \\
                 & (0.005)     \\
                       &           \\
    \textbf{Controls}            &           \\
          {}   Time of day           & Yes  \\
      {}  Day of Week           & Yes \\
      {}  Week of Year           & Yes \\
      {}  Year           & Yes \\
    {} Chief Complaint    & Yes       \\
    {} Physician          & Yes       \\ \bottomrule
    \end{tabular}
    \caption{Racial bias in the predicted fatigue score with separation on Hispanic by the reported languages. The regression only includes patients using English or Spanish, which results in 32,827 encounters (96\% of total encounter of the heldout set). 3260 of Hispanic patients reported using Spanish, and 4634 of Hispanic patients reported using English.}
    \label{tab:tiredness-race-language}
\end{table*}

\section{Generated Notes Have Higher Fatigue Scores}
We genearte the History of Present Illness (HPI) section of the notes with our finetuned GPT-2 model. 
For each note, we truncate the notes at the HPI section heading and generate the HPI section with greedy and sample decoding.
We then obtain the fatigue score of the generated and original HPI sections with the fatigue predictor using the same set of features.
Table~\ref{tab:gen-hpi} shows that the predicted fatigue score is higher for the generated notes.
We observe that generated notes have higher scores on log perplexity for both greedy and sample decoding, and the fraction of anger words increases for sample decoding, which is the standard practice for recent large language models, e.g., GPT-4.\footnote{The default top\_p value is 1 in the OpenAI API documentation (\url{https://platform.openai.com/docs/api-reference}). Results are robust with different top\_p.}

\begin{table}[]
    \centering
    \begin{tabular}{@{}lccc@{}}
    \toprule
    & Predicted Fatigue & Log Perplexity & Fraction of \\
     & Deviation from the Mean &  &Anger Words   \\ \midrule
    \textbf{Decoding method}&\\
    {} Greedy          & 189\%  & 0.5983& 0.03\%  \\
    {} Sampled         & 8\% & 1.3335&0.12\% \\ \midrule
    Original HPI    & -66\%  & 1.6141& 0.09\% \\ \bottomrule
    \end{tabular}
    \caption{Fatigue score of generated notes with different decoding methods. Greedy decoding generates notes with higher fatigue scores than sampled decoding. Both generated notes have higher fatigue scores than the original HPI notes. Results are based on 1471 sampled notes from the heldout set.
    }
    \label{tab:gen-hpi}
\end{table}

\section{Feature importance}

Table~\ref{tab:correlation_full} shows the full table of feature importance based on Pearson’s correlation
with the high- vs. low-workload indicator, the yield of testing indicator, and the non-White indicator.

\begin{table*}[]
\centering
\begin{tabular}{@{}lc@{}}
\toprule
Feature                                    & Pearson’s Correlation              \\ \midrule
\textbf{Note Feature}                      & \multicolumn{1}{l}{} \\
{} note length                                & -0.014**             \\
{} note predictability                             & -0.092***            \\
{} fraction of stopwords                      & -0.009               \\
{} fraction of medical words         & 0.006                     \\\\
\textbf{LIWC Pronoun}                      & \multicolumn{1}{l}{} \\
{} fraction of pronoun                        & -0.017***           \\
{} {} fraction of first person singular pronouns & -0.07***             \\
{} {} fraction of first person plural pronouns   & 0.017***            \\
{} {} fraction of second person pronouns         & -0.001              \\
{} {} fraction of third person singular pronouns & 0.008                 \\
{} {} fraction of third person plural pronouns   & 0.006                 \\
{} {} fraction of impersonal pronouns   & -0.049***          \\\\
\textbf{LIWC Affect}                       & \multicolumn{1}{l}{}  \\
{} fraction of affective processes            & -0.005              \\
{} {} fraction of positive emotions              & -0.003               \\
{} {} fraction of negative emotions              & -0.006               \\
{} {} fraction of anxiety                        & -0.001               \\
{} {} fraction of anger                          & 0.025***            \\
{} {} fraction of sadness               & 0.002                          \\\\
\textbf{LIWC Congnitive}                   & \multicolumn{1}{l}{}  \\
{} fraction of cognitive processes            & 0.023***             \\
{} {} fraction of insight                        & -0.09***             \\
{} {} fraction of causation                      & 0.025***              \\
{} {} fraction of discrepancy                    & 0.026***             \\
{} {} fraction of tentative                      & 0.027***            \\
{} {} fraction of certainty                      & 0.075***             \\
{} {} fraction of inhibition                     & 0.039***             \\
{} {} fraction of inclusive                      & 0.005                 \\
{} {} fraction of exclusive             & 0.041***      \\ \\
\textbf{Readability}                       & \multicolumn{1}{l}{}  \\
{} {} Flesch Kincaid grade                       & -0.05***            \\\midrule          
\end{tabular}
\caption{Full table of feature importance based on Pearson's correlation with the high- vs. low-workload indicator. 
    We use the name of the corresponding LIWC category to represent the name of a set of lexical words, so ``fraction of inclusive'' should be interpreted as ``fraction of the inclusive category''.}
\label{tab:correlation_full}
\end{table*}

\end{document}